\documentclass{IOS-Book-Article}

\usepackage{mathptmx}
\usepackage{soul}\setuldepth{article}

\usepackage{amsmath}
\usepackage{graphicx}
\usepackage[T1]{fontenc}   
\usepackage{ccicons}          
\usepackage{paralist}
\usepackage{hyperref}
\usepackage[nolist]{acronym}
\usepackage{color}

\usepackage[table,xcdraw]{xcolor}

\def\hb{\hbox to 11.5 cm{}}

\newcommand{\quotes}[1]{``#1''}

\begin{document}

\pagestyle{headings}
\def\thepage{}

\begin{frontmatter}              

\title{A Survey on Methods and Metrics for the Assessment of Explainability under the Proposed AI Act}


\author[A]{\fnms{Francesco} \snm{Sovrano}}
and
\author[B]{\fnms{Salvatore} \snm{Sapienza}}
and
\author[B]{\fnms{Monica} \snm{Palmirani}}
and
\author[A]{\fnms{Fabio} \snm{Vitali}}

\address[A]{University of Bologna, DISI}
\address[B]{University of Bologna, CIRSFID}

\begin{abstract}
This study discusses the interplay between metrics used to measure the explainability of the AI systems and the proposed EU Artificial Intelligence Act. 
A standardisation process is ongoing: several entities (e.g. ISO) and scholars are discussing how to design systems that are compliant with the forthcoming Act and explainability metrics play a significant role. This study identifies the requirements that such a metric should possess to ease compliance with the AI Act. It does so according to an interdisciplinary approach, i.e. by departing from the philosophical concept of explainability and discussing some metrics proposed by scholars and standardisation entities through the lenses of the explainability obligations set by the proposed AI Act. Our analysis proposes that metrics to measure the kind of explainability endorsed by the proposed AI Act shall be \textit{risk-focused}, \textit{model-agnostic}, \textit{goal-aware}, \textit{intelligible \& accessible}. This is why we discuss the extent to which these requirements are met by the metrics currently under discussion. 

\end{abstract}

\end{frontmatter}

\begin{acronym}
	\acro{EU}{European Union}
	\acro{ADM}{Automated Decision-Making system}
	\acro{AI-HLEG}{High-Level Expert Group on Artificial Intelligence}
	\acro{AI}{Artificial Intelligence}
	\acro{XAI}{eXplainable AI}
	\acro{YAI}{explanatorY AI}
	\acro{HCI}{Human-Computer Interaction}
	\acro{GDPR}{General Data Protection Regulation}
	\acro{RDF}{Resource Description Framework}
	\acro{CEM}{Contrastive Explanations Method}
	\acro{KB}{Knowledge Base}
	\acro{DPO}{Data Protection Officer}
	\acro{DPIA}{Data Protection Impact Assessment}
	\acro{AIA}{Artificial Intelligence Act}
	\acro{DoX}{Degree of eXplainability}
	\acro{WeDoX}{Weighted Degree of eXplainability}
\end{acronym}

\section{Introduction}
The ability and need of humans to explain has been studied for centuries, initially in philosophy and more recently also in all those sciences aiming at a better understanding of (human) intelligence.
Measuring the degree of explainability of AI systems has become relevant in the light of research progress in the \ac{XAI} field, the proposal for an EU Regulation on Artificial Intelligence, and ongoing standardisation initiatives that will translate these technical advancements in a \textit{de facto} regulatory standard for AI systems. 
To date, standardisation entities have proposed white papers and preliminary documents showing their progress
, among them we mention:
\begin{inparadesc}
    \item the European Telecommunications Standards Institute (ETSI),
    \item the CEN-CENELEC,
    \item and ISO/IEC TR 24028:2020(E), stating that '[i]t is important also to consider the measurement of the quality of explanations' and provides for details on the key measurements (i.e. continuity, consistency, selectivity; paras 9.3.6, 9.3.7).
\end{inparadesc}

Considering that, since ISO/IEC TR 24028:2020(E), the literature has started to propose new metrics and mechanisms, with this work we study and categorise the existing approaches to quantitatively assess the quality of explainability in Machine Learning and AI. 
We do so through the lenses of law and philosophy, not just computer science.
This last characteristic is certainly our main contribution to the literature of \ac{XAI} and Law, and we believe it may foster future research to embrace an interdisciplinary approach less timidly, for the sake of a better conformity to existing (and new) regulations in the EU panorama.

This paper is structured as follows. 
In Section \ref{sec:related_work} and \ref{sec:methodology} we present the research background and the methodology of this paper.
Then in Section \ref{sec:philosophy}, \ref{sec:explainability_desiderata} and \ref{sec:law}, we explore the definitions and properties of explainability in philosophy and in the proposed AI Act. 
Finally, in Section \ref{sec:metrics} and \ref{sec:final_remarks} we perform an analysis of the existing quantitative metrics of explainability, discussing our findings and future research. 

\section{Related Work} \label{sec:related_work}


In XAI's literature there are many interesting surveys on explainability techniques \cite{guidotti2018survey,adadi2018peeking,arrieta2020explainable,zhou2021evaluating}, classifying algorithms on different dimensions to help researchers in finding the more appropriate ones for their own work.
Practically, all these surveys focus on a classification of the mechanisms to achieve explainability rather than how to measure the quality of it, and we believe our work can help in this latter.

For example \cite{guidotti2018survey} classify XAI methods with respect to the notion of explanation and the type of black-box system. The identified characteristics are respectively the level-of-detail of explainability (from high to low: global logic, local decision logic, model properties) and the level of interpretability of the original model.
Similarly to \cite{guidotti2018survey}, also \cite{adadi2018peeking} study XAI considering interpretability and level-of-detail.

On the other hand, \cite{zhou2021evaluating} focus specifically on the metrics to quantify the quality of explanation methods, classifying them according to the properties they can measure and the format of explanations (model-based, attribution-based, example-based) they support. 
More precisely, 
\cite{zhou2021evaluating} narrow down the survey to the functionality-grounded metrics, proposing for them a new taxonomy including interpretability (in terms of clarity, broadness, and parsimony) and fidelity (as completeness, and soundness).

Among all the identified surveys, \cite{zhou2021evaluating} is certainly the closest to our work, in terms of focus of the survey.
The main distinction between our work and \cite{zhou2021evaluating} is probably the assumption we do that 
multiple definitions of explainability exist, each one possibly requiring its own type of metrics.
Furthermore, differently from \cite{zhou2021evaluating}, we analyse explainability metrics on their ability to meet the requirements set by the AI Act. 

\section{Methodology} \label{sec:methodology}
We performed an exploratory literature review of existing metrics to measure the explainability of AI-related explanations, together with a qualitative legal analysis of the explainability requirements to understand the alignment of the identified metrics to the expectations of the proposed AI Act.
To do so, we collected all the papers cited in \cite{zhou2021evaluating}, re-classifying them.
Then we integrated with further works identified through an in depth keyword-based research\footnote{The main keywords we used were \quotes{degree of explainability}, \quotes{explainability metrics}, \quotes{explainability measures}, and \quotes{evaluation metrics for contrastive explanations}.} on Google Scholars, Scopus, and Web Of Science.
On the other hand, the legal analysis was carried out on the proposed Artificial Intelligence Act. 
Considering the lack of case law and the paucity of studies on this novel piece of legislation, a literal assessment of its provisions has been preferred to more critical analysis based on previous enquiries.

\begin{table}[!tb]
\caption{\textbf{Definitions of \textit{explanation} and \textit{explainable information}} for each theory of explanations.} \label{tab:definitions}
\makebox[\linewidth]{
\begin{tabular}{|p{0.25\linewidth}|p{0.7\linewidth}|p{0.35\linewidth}|}
\hline
\rowcolor[HTML]{C0C0C0} 
\textbf{Theory}                            & \textbf{Def. of Explanation}                                                                                                                                                                                                                                                                           & \textbf{Def. of Explainable Information}                                                         \\ \hline
Causal Realism \cite{salmon1984scientific}                   & It is a description of causality, as chains of causes and effects.                                                                                                                                                                                                                            & It can fully describe causality.                                                        \\ \hline
Constructive Empiricism \cite{van1980scientific}          & It is contrastive information answering WHY questions, allowing one to calculate the probability of a particular event relative to a set of (possibly subjective) background assumptions.                                                                                                     & It provides answers to contrastive WHY questions.                                       \\ \hline
Ordinary Language Philosophy \cite{achinstein1983nature}     & Explaining is pragmatically answering to (not just WHY) questions, with the explicit intent of producing understanding.                                                                                                                                                                                       & It can be used to pertinently answer questions about relevant aspects, in an illocutionary way. \\ \hline
Cognitive Science \cite{holland1989induction}                & Explaining is a process triggered as response to predictive failures and it is about providing information to fix that failures in a mental model (sometimes intended as a hierarchy of rules).                                                                                               & It can fix failures in mental models.                                                   \\ \hline
Naturalism and Scientific Realism \cite{sellars1963philosophy} & Explaining is an iterative process of confirmation of truth based on inference to the best explanation. An explanation increases understanding, not simply by being the correct answer to a particular question, but by increasing the coherence of an entire belief system (e.g. a subject). & It can be used to increase understanding, i.e. by answering to particular questions.    \\ \hline
\end{tabular}
}
\end{table}

\section{Definitions of Explainability} \label{sec:philosophy}
Considering the definition of \quotes{explainability} as \quotes{the potential of information to be used for explaining}, we envisage that a proper understanding of how to measure explainability must pass through a thorough definition of what constitutes an explanation and the act of explaining.

In 1948 Hempel and Oppenheim published their \quotes{Studies in the Logic of Explanation} \cite{hempel1948studies}, giving birth to what it is considered the first theory of explanation, the deductive-nomological model.
After that date, many attempts followed to amend, extend or replace this first model, which is considered fatally flawed \cite{bromberger1966questions,salmon1984scientific}.
This gave birth to several competing and more contemporary theories of explanations \cite{mayes2005theories}:
\begin{inparaenum}[i)]
    \item Causal Realism,
    \item Constructive Empiricism, 
    \item Ordinary Language Philosophy,
    \item Cognitive Science,
    \item Naturalism and Scientific Realism.
\end{inparaenum}
A summary of these definitions is shown in Table \ref{tab:definitions}.

Interestingly, each one of these theories devises different definitions of \quotes{explanation}. 
If we look at their specific characteristics we may find that all but \textit{Causal Realism} are pragmatic. 
On the other hand, \textit{Causal Realism} and \textit{Constructive Empiricism} are rooted on causality, while the others not \footnote{They study the act of explaining as an iterative process involving broader forms of question answering}.
Nonetheless, \textit{Cognitive Science} and \textit{Scientific Realism} are more focused on the effects that an explanation has on the explainee (the recipient of the explanation).

Importantly, with the present letter, we assert that whenever explaining is considered to be a pragmatic act, explainability differs from explaining.
In fact, pragmatism in this sense is achieved when the explanation is tailored to the specific user, so that the same explainable information can be presented and re-elaborated differently across users. 
It follows that for each philosophical tradition, but Causal Realism, we have a definition of \quotes{explainable information} that slightly differs from that of \quotes{explanation}, as shown in Table \ref{tab:definitions}.

\section{Explainability Desiderata} \label{sec:explainability_desiderata}

In philosophy, the most important work about the central criteria of adequacy of \textit{explainable information} is likely to be Carnap's \cite{leitgeb2021carnap}. 
Even though Carnap studies the concept of \textit{explication} rather than that of \textit{explainable information}, we assert that they share a common ground making his criteria fitting in both cases.
In fact, \textit{explication} in Carnap’s sense is the replacement of a somewhat unclear and inexact concept (the explicandum) by a new, clearer, and more exact concept called explicatum, and that is exactly what information does when made explainable.

Carnap’s central criteria of explication adequacy are \cite{leitgeb2021carnap}: \textit{similarity}, \textit{exactness} and \textit{fruitfulness}\footnote{Carnap also discussed another desideratum, \textit{simplicity}, but this criterion is presented as being subordinate to the others.}.
\textit{Similarity} means that the explicatum should be similar to the explicandum, in the sense that at least many of its intended uses, brought out in the clarification step, are preserved in the explicatum.
On the other hand, \textit{Exactness} means that the explication should, where possible, be embedded in some sufficiently clear and exact linguistic framework.
While \textit{Fruitfulness} means that the explicatum should be used in a high number of other \textit{good} explanations (the more, the better). 

Carnap's adequacy criteria seem to be transversal to all the identified definitions of explainability, possessing preliminary characteristics for any piece of information to be considered properly explainable.
Therefore, our interpretation of Carnap’s criteria in terms of measurements is the following.
\begin{itemize}
    \item \textit{Similarity} is about measuring how much \textit{similar} the given information is to the explanandum. This can be estimated by counting the number of \textit{relevant aspects} covered by information and the \textit{amount of details} it can provide.
    \item \textit{Exactness} is about measuring how clear the given information is, in terms of pertinence and syntax, regardless its truth. Differently from Carnap, our understanding of \textit{exactness} is broader than that of adherence to standards of formal concept formation \cite{brun2016explication}.
    \item \textit{Fruitfulness} is about measuring how much a given piece of information is going to be used in the generation of explanations. Consequently, each one of the explainability definitions may define \textit{fruitfulness} differently.
\end{itemize}
Importantly, the property of \textit{truthfulness} (being different from \textit{exactness}) is not explicitly mentioned in Carnap's desiderata.
That is to say that explainability and \textit{truthfulness} are complementary, but different, as discussed also by \cite{hilton1996mental}. In fact an explanation is such regardless its truth (wrong but high-quality explanations exist, especially in science). Vice-versa, highly correct information can be very poorly explainable.

\section{Explainability Obligations in the Proposed AI Act} \label{sec:law}

Following the EU Commission's Proposal for an \ac{AIA}, it is now time to discuss how explainability is connected to the novel obligations introduced by the Act. 
In fact, considering the nature and the characteristics of the requirements posed by the \ac{AIA}, it is worth questioning how explainability metrics could be designed to fulfil the necessities of all the entities whose behaviour will be regulated by the \ac{AIA}. 

The discussion towards \quotes{explainability and law} has departed from the contested existence of a right to explanation in the \ac{GDPR}
\cite{wachter2017right, selbst2018meaningful} 
to embrace contract, tort, banking law \cite{hacker2021varieties}, and judicial proceedings \cite{ebers2020regulating}.
Differently from other domains, the \ac{AIA} is specific to AI systems and requires an \textit{ad hoc} discussion rather than the framing of these systems in the discussion of other legal domains. 
This is because AI technologies are not placed within an existing legal framework (e.g. banking), but the whole legal framework (i.e. the \ac{AIA}) is built around AI technologies. 
However, the previous discussion focusing on other legal regimes constitutes a valuable background for our research and thus it contributes to our discussion. 
The interpretations proposed by recent commentators \cite{ebers2020regulating, hacker2021varieties} identify several nuances of algorithmic transparency. 
Our focus, however, shall be confined to the interaction between the nuance of explainability and obligations emerging from the \ac{AIA} already identified by these early commentators.

As regards the \ac{GDPR}, scholars have extensively discussed whether or not the right to receive an explanation for \quotes{solely automated decision-making} processes exists in the \ac{GDPR}
\footnote{
    Regardless of the answer, the data controller has an obligation to provide \quotes{meaningful information about the logic involved} in the automated decision. See art. 13(2)(f), art. 14(2)(g), art. 15(1)(h).
}. 
Then, the discussion identified a \quotes{technical} necessity of explainability, that is necessary to improve the accuracy of the model. 
In legal terms, it is echoed by the \quotes{protective} transparency that is needed to minimise risks and comply with certain legal regimes (tort law and contractual obligations). 
As with data protection law, these varieties are instrumental to improve a product and protect its users or the persons affected by the system from damages.
If explainability is often instrumental to achieve some legislative goals, it is likely that it could be meant to foster certain regulatory purposes also under the \ac{AIA}. 
From the joint reading of a series of provisions, it will be argued that explainability in the \ac{AIA} is both \textit{user-empowering} and \textit{compliance-oriented}: on the one hand, it serves to enable users of the AI system to use it correctly; on the other hand, it helps to verify adequacy to the many obligations set by the \ac{AIA}.  

Recital 47 and art. 13(1) state that high-risk AI systems shall be designed and developed in such a way that their operation is comprehensible by the users. 
They should be able a) to interpret the system's output and b) to use it in an appropriate manner. 
This is a form of \textit{user-empowering} explainability.
Then, the second part of Art. 13 specifies that \quotes{an appropriate type and degree of transparency shall be ensured, with a view to \textit{achieving compliance} (emphasis added) with the relevant obligations of the user and of the provider [...]}. 
In our reading, this provision specifies that this explainability obligations (i.e. transparent design and development of high-risk AI systems) is \textit{compliance-oriented}. 
The twofold goal of art. 13(1) is then echoed by other provisions. As regards the user-empowering interpretation, art. 14(4)(c) relates explainability to \quotes{human oversight} design obligations. These measures should enable the individual supervising the AI system to correctly interpret its output. Moreover, this interpretation shall put him or her in the position to decide whether it might be the case to \quotes{disregard, override or revers the output}, art. 14(4)(d). 

The compliance-oriented explainability interpretation becomes evident in the technical documentation to be provided according to Article 11. 
Compliance is based on a presumption of safety if the system is designed according to technical standards (Art. 40) to which adherence is documented, whereas third-party assessment appears only post-market or on specific sectors (see Chapter IV).
The contents of the dossier are those detailed by Annex IV. 
\textit{Inter alia}, Annex IV(2)(b) include \quotes{the design specifications of the system, namely the general logic of the AI system and of the algorithms} among the information to be provided to show compliance with the \ac{AIA} before placing the AI system in the market.
Hence, the system should be explainable in a manner that allows an evaluation of conformity by the provider in the first instance and, when necessary, by post-market monitoring authorities. 
Since the general approach taken by the proposed \ac{AIA} is a risk-reduction mechanism (Recital 5), this form of explainability is ultimately meant to contribute to minimising the level of potential harmfulness of the system.

User-empowering and compliance oriented explainability overlap in art. 29(4).
When a risk is likely to arise, the user shall suspend the use of the system and inform the provider or the distributor. 
This provision entails the capability of understanding the working of the system (real-time) and making previsions on its output. 
Suspending in the case of likely risk is the overlapping between the two nuances of explainability: the user is empowered to stop the AI system to avoid contradicting the rationale behind the \ac{AIA}, i.e. risk-minimisation. 

Once clarified the existence of explainability obligations and their extent, let us discuss the requirements that metrics should have to ease compliance with the \ac{AIA}. 
Let us remind that, under the proposal, adopting a standard means certifying the degree of explainability of a given AI system. 
Therefore, metrics become useful in the course of the  standardisation process: 
\begin{inparaenum}[i)]
    \item \textit{ex ante}, when defining the explainability measures adopted by the standard; 
    \item \textit{ex post}, when verifying in practice the adoption of a standard. 
\end{inparaenum}

From these premises it follows that, in the light of the purposes of the \ac{AIA}, any explainability metric should be at minimum:
\begin{inparaenum}[i)]
    \item \textit{Risk-focused},
    \item \textit{Model-agnostic},
    \item \textit{Goal-Aware},
    \item and \textit{Intelligible \& accessible}.
\end{inparaenum}

\textit{Risk-focused} means that the metric should be functional to measure the extent to which the explanations provided by the system allows for an assessment of the risks to the fundamental rights and freedoms of the persons affected by the system's output. 
This is necessary to ensure both user-enabling (e.g. art. 29) and compliance-oriented (Annex IV) explainability.
While \textit{Model-agnostic} means that the metric should be appropriate to all the AI systems regulated by the \ac{AIA}\footnote{Annex I provides a list of the AI techniques and approaches that fall within the remit of the Regulation.}.

\textit{Goal-aware} means that the metric should be flexible towards the different needs of the potential explainees (i.e. AI system providers and users, standardisation entities, etc.)\footnote{Since it might be hard to determine \textit{ex ante} the nature, the purpose, and the expertise of the explainee, the metrics should consider the highest possible number of potential explainees.} and applicable in all the high-risk AI applications listed in Annex III.
While \textit{Intelligible \& accessible} means that if information on the metrics is not accessible (e.g. due to intellectual property reasons) or the results of a metric are not reproducible (e.g. due to a subjective evaluation), explainees will confront with a situation of uncertainty, as an \textit{ignotum per ignotius}. 
This would contradict the risk minimisation principle.

\section{Discussing Existing Quantitative Measures of Explainability} \label{sec:metrics}

In this section we identify some pros and cons of existing metrics (and measures) to quantitatively estimate the degree of explainability of information, with the aim of understanding their range of applicability across different needs and interpretations of explainability.
We do it by performing a qualitative classification of these measures based on Carnap's desiderata, the theories of explanation presented in Section \ref{sec:philosophy} and the main principles identified in Section \ref{sec:law}.

More precisely, in Table \ref{tab:literature_comparison} we classified the metrics on the following dimensions:
\begin{inparadesc}
    \item the \textit{format of information} supported by the metric (i.e. rule-based, example-based, natural language text, etc.);
    \item the \textit{supporting theory of the metric} (i.e. cognitive science, constructive empiricism, etc.);
    \item \textit{subjectivity} (whether the metric requires evaluations given by humans subjects);
    \item the \textit{covered criteria of adequacy}.
\end{inparadesc}
Then, in Table \ref{tab:philosophy_vs_law} we aligned the \textit{supporting theories} (hence also the metrics) to the properties identified with the analysis of the AI Act carried out in Section \ref{sec:law}.


Doing so, we considered only a part of the dimensions adopted by \cite{zhou2021evaluating}. 
More precisely, we kept \textit{clarity}, \textit{broadness} and \textit{completeness}, aligning the first two to Carnap's \textit{exactness} and the latter to \textit{similarity}.
In fact, we deemed \textit{soundness} to be as \textit{truthfulness}, a complementary characteristic to explainability and not a characteristic of explainability, as discussed in Section \ref{sec:explainability_desiderata}.
While \textit{broadness} and \textit{parsimony} were considered as characteristics to achieve pragmatic explanations rather than properties of explainability.

Furthermore, differently from ISO/IEC TR 24028:2020(E) we did not focus on metrics specific to ex-post \textit{feature attribution} explanations, so we selected methods possibly applicable also on ex-ante or more generic types of explanations.

\begin{table}[!tb]
\caption{\textbf{Comparison of different explainability metrics.} The column \quotes{Metric} points to reference papers, while column \quotes{Name} points to the names used by the authors of the metric to describe it. Elements in bold are column-wise, indicating the best values.}\label{tab:literature_comparison}
\makebox[\linewidth]{
\begin{tabular}{|p{.06\linewidth}|p{.19\linewidth}|p{.23\linewidth}|p{.07\linewidth}|p{.15\linewidth}|p{.25\linewidth}|}
\hline
\rowcolor[HTML]{C0C0C0} 
\textbf{Metric}                                                                                            & \textbf{Information Format}                                                            & \textbf{Supporting Theory}                                                                     & \textbf{Subject - based} & \textbf{Covered Criteria}                                                                          & \textbf{Name}                                                                                                                                          \\ \hline
\cite{villone2020comparative}                                                    & Rule-based                                                                    & Causal Realism                                                                        & \textbf{No}   & \begin{tabular}[c]{@{}l@{}}Similarity, \\ Fruitfulness\end{tabular}                        & \begin{tabular}[c]{@{}l@{}}Fidelity, \\ Completeness\end{tabular}                                                                             \\ \hline
\cite{nguyen2020quantitative}                                                    & Feature Attribution & Causal Realism & \textbf{No}   & \begin{tabular}[c]{@{}l@{}}Similarity, \\ Fruitfulness\end{tabular}                        & \begin{tabular}[c]{@{}l@{}}Monotonicity, \\ Non-sensitivity, \\ Effective Complexity\end{tabular} \\ \hline  
\cite{lakkaraju2017interpretable}                                                & Rule-based                                                                    & Causal Realism                                                                        & \textbf{No}   & \textbf{\begin{tabular}[c]{@{}l@{}}Similarity, \\ Exactness, \\ Fruitfulness\end{tabular}} & \begin{tabular}[c]{@{}l@{}}Fidelity, Unambiguity, \\ Interpretability, \\ Interactivity\end{tabular}                                       \\ \hline
\cite{holzinger2020measuring}                                                    & \textbf{All}                                                                  & \begin{tabular}[c]{@{}l@{}}Causal Realism, \\ Cognitive Science, \\ Scientific Realism\end{tabular}          & Yes           & \begin{tabular}[c]{@{}l@{}}Exactness, \\ Fruitfulness\end{tabular}                         & Causability                                                                                                                                   \\ \hline
\cite{hoffman2018metrics}                                                        & \textbf{All}                                                                  & \begin{tabular}[c]{@{}l@{}}Cognitive Science, \\ Scientific Realism\end{tabular}                                                                     & Yes           & \begin{tabular}[c]{@{}l@{}}Exactness, \\ Fruitfulness\end{tabular}                         & \begin{tabular}[c]{@{}l@{}}Satisfaction, Trust, \\ Mental Models, \\ Curiosity, Performance\end{tabular}                                \\ \hline
\cite{keane2021if} & Example-based & \begin{tabular}[c]{@{}l@{}}Constructive Empiricism\end{tabular} & \textbf{No}           & \begin{tabular}[c]{@{}l@{}}Exactness\end{tabular}                        & \begin{tabular}[c]{@{}l@{}}Proximity, Sparsity, \\ Adequacy (Coverage) \end{tabular}                                               \\ \hline
\cite{nguyen2020quantitative}                                                    & Example-based & Constructive Empiricism & \textbf{No}   & \begin{tabular}[c]{@{}l@{}}Similarity, \\ Fruitfulness\end{tabular}                        & \begin{tabular}[c]{@{}l@{}}Non-Representativeness, \\ Diversity\end{tabular} \\ \hline
\cite{sovrano2021metric}                                                         & Natural Language Text                                                         & Ordinary Language                                                          & \textbf{No}   & \textbf{\begin{tabular}[c]{@{}l@{}}Similarity, \\ Exactness, \\ Fruitfulness\end{tabular}} & \begin{tabular}[c]{@{}l@{}}Aspects Coverage, \\ Degree of Explainability\end{tabular}                                                         \\ \hline
\end{tabular}
}
\end{table}

As shown in Table \ref{tab:literature_comparison}, we were able to find at least one example of metric for each supporting philosophical theory
, with a majority of metrics focused on Causal Realism and Cognitive Science.
What is common to all the metrics based on Cognitive Science is that they require humans subjects for performing the measurement, therefore they tend to be more expensive than the others, at least in terms of human effort.
Furthermore, the metrics proposing heuristics to measure all Carnap's desiderata are just two, one for Causal Realism \cite{lakkaraju2017interpretable} and the other for Ordinary Language Philosophy \cite{sovrano2021metric}. Interestingly, \cite{lakkaraju2017interpretable} evaluates the three desiderata separately, while \cite{sovrano2021metric} propose a single metric combining all of them.
Finally, the results shown in Table \ref{tab:philosophy_vs_law} indicate that the metrics supported by both Causal Realism and Constructive Empiricism might struggle at being model-agnostic and goal-aware, this probably limits their applicability to very specific contexts.

\begin{table}[!tb]
\caption{\textbf{Explainability definitions alignment} to the properties identified in Section \ref{sec:law}.} \label{tab:philosophy_vs_law}
\makebox[\linewidth]{
\begin{tabular}{l|p{0.2\linewidth}|p{0.25\linewidth}|p{0.2\linewidth}|p{0.3\linewidth}|}
\cline{2-5}
\cellcolor[HTML]{FFFFFF}\textbf{}                                               & \cellcolor[HTML]{C0C0C0}\textbf{Risk-Focused} & \cellcolor[HTML]{C0C0C0}\textbf{Model-Agnostic}                              & \cellcolor[HTML]{C0C0C0}\textbf{Goal-Aware}   & \cellcolor[HTML]{C0C0C0}\textbf{Intelligible \& Accessible}                                                                                      \\ \hline
\multicolumn{1}{|p{0.2\linewidth}|}{\cellcolor[HTML]{EFEFEF}Causal Realism}                    & Yes, if understanding risks implies understanding causality                        & Not available yet                                                                & No, it's not pragmatic and it considers only goals related to causality & Yes, it can be                                                                                                                              \\ \hline
\multicolumn{1}{|p{0.2\linewidth}|}{\cellcolor[HTML]{EFEFEF}Constructive Empiricism}           & Yes, if explaining risks is about answering WHY questions                         & Not available yet                                                                & No, it focuses only on WHY questions             & Yes, it can be                                                                                                                              \\ \hline
\multicolumn{1}{|p{0.2\linewidth}|}{\cellcolor[HTML]{EFEFEF}Ordinary Language Philosophy}      & Yes, it can be                                           & Maybe. Only if all the explanations can be represented in a natural language & Yes                                              & Yes, it can be                                                                                                                              \\ \hline
\multicolumn{1}{|p{0.2\linewidth}|}{\cellcolor[HTML]{EFEFEF}Cognitive Science}                 & Yes, it can be                                           & Yes, the evaluation is subject-based                                         & Yes                                              & Unlikely. All the subject-based metrics may be very expensive and hard to reproduce, this makes them less accessible \\ \hline
\multicolumn{1}{|p{0.2\linewidth}|}{\cellcolor[HTML]{EFEFEF}Naturalism and Scientific Realism} & Yes, it can be                                           & Yes, the evaluation is subject-based                                         & Yes                                              & Unlikely. It relies on (usually) expensive subject-based metrics                                                                                    \\ \hline
\end{tabular}
}
\end{table}

\section{Final Remarks} \label{sec:final_remarks}
With this work we proposed an interdisciplinary analysis of explainability metrics in Artificial Intelligence. 
More specifically, through the lens of the obligations enshrined by the proposed Act, we identified that explainability metrics should be \textit{risk-focused}, \textit{model-agnostic}, \textit{goal-aware}, \textit{intelligible \& accessible}. 
We found that these characteristics pose some constraints on the scope of explainability metrics, suggesting that different metrics may be complementary, serving different roles, depending on the context.
In fact, as shown in Table \ref{tab:philosophy_vs_law}, while the majority of \textit{supporting theories} have the potential to result in \textit{risk-focused} metrics, some of them might have important issues with \textit{goal-awareness}, \textit{intelligibility} and \textit{accessibility}.

Nonetheless, our analysis of these metrics was qualitative and not quantitative.
In fact, all of the considered metrics were tested by their authors on very specific applications and technologies, raising the issue of whether they can be seemingly effective under different implementation scenarios.
Hence, we envisage that a more quantitative analysis should be carried on, perhaps by defining a proper benchmark on which metrics can be thoroughly evaluated from a legal perspective.

Therefore, we believe that more academic contributions and new benchmarks for quantitative legal analysis are needed,
to better understand the pros and cons of existing technologies, for any standardisation process to be finalised and effectively deployed in the EU panorama.
For example, considering the current level of discussion and that our findings might be subject to change due to the institutional debate about the Proposal, further research is needed at least to consolidate the interpretation of the Act in the light of its future changes. 

\bibliographystyle{vancouver}
\bibliography{biblio}

\end{document}